\documentclass[sigconf]{acmart}
\renewcommand\footnotetextcopyrightpermission[1]{} % removes footnote with conference information in first column
\usepackage{multirow}
\usepackage{arydshln}
\AtBeginDocument{%
  \providecommand\BibTeX{{%
    \normalfont B\kern-0.5em{\scshape i\kern-0.25em b}\kern-0.8em\TeX}}}

\begin{document}

\title{Situational Perception Guided Image Matting}

\author{Bo Xu\textsuperscript{1}, Jiake Xie\textsuperscript{2}, Han Huang\textsuperscript{1}, Ziwen Li\textsuperscript{1}, Cheng Lu\textsuperscript{3}, Yong Tang\textsuperscript{2} and Yandong Guo\textsuperscript{1,*}\\
\textsuperscript{1}OPPO Research Institute, \textsuperscript{2}PicUp.AI, \textsuperscript{3}Xmotors}
\newcommand{\etal}{\textit{et al}.}




\begin{abstract}
Most automatic matting methods try to separate the salient foreground from the background. However, the insufficient quantity and subjective bias of the current existing matting datasets make it difficult to fully explore the semantic association between object-to-object and object-to-environment in a given image. In this paper, we propose a Situational Perception Guided Image Matting (SPG-IM) method that mitigates subjective bias of matting annotations and captures sufficient situational perception information for better global saliency distilled from the visual-to-textual task. SPG-IM can better associate inter-objects and object-to-environment saliency, and compensate the subjective nature of image matting and its expensive annotation. We also introduce a textual Semantic Transformation (TST) module that can effectively transform and integrate the semantic feature stream to guide the visual representations. In addition, an Adaptive Focal Transformation (AFT) Refinement Network is proposed to adaptively switch multi-scale receptive fields and focal points to enhance both global and local details. Extensive experiments demonstrate the effectiveness of situational perception guidance from the visual-to-textual tasks on image matting, and our model outperforms the state-of-the-art methods. We also analyze the significance of different components in our model. The code will be released soon.
\end{abstract}



\begin{teaserfigure}
  \includegraphics[width=\textwidth]{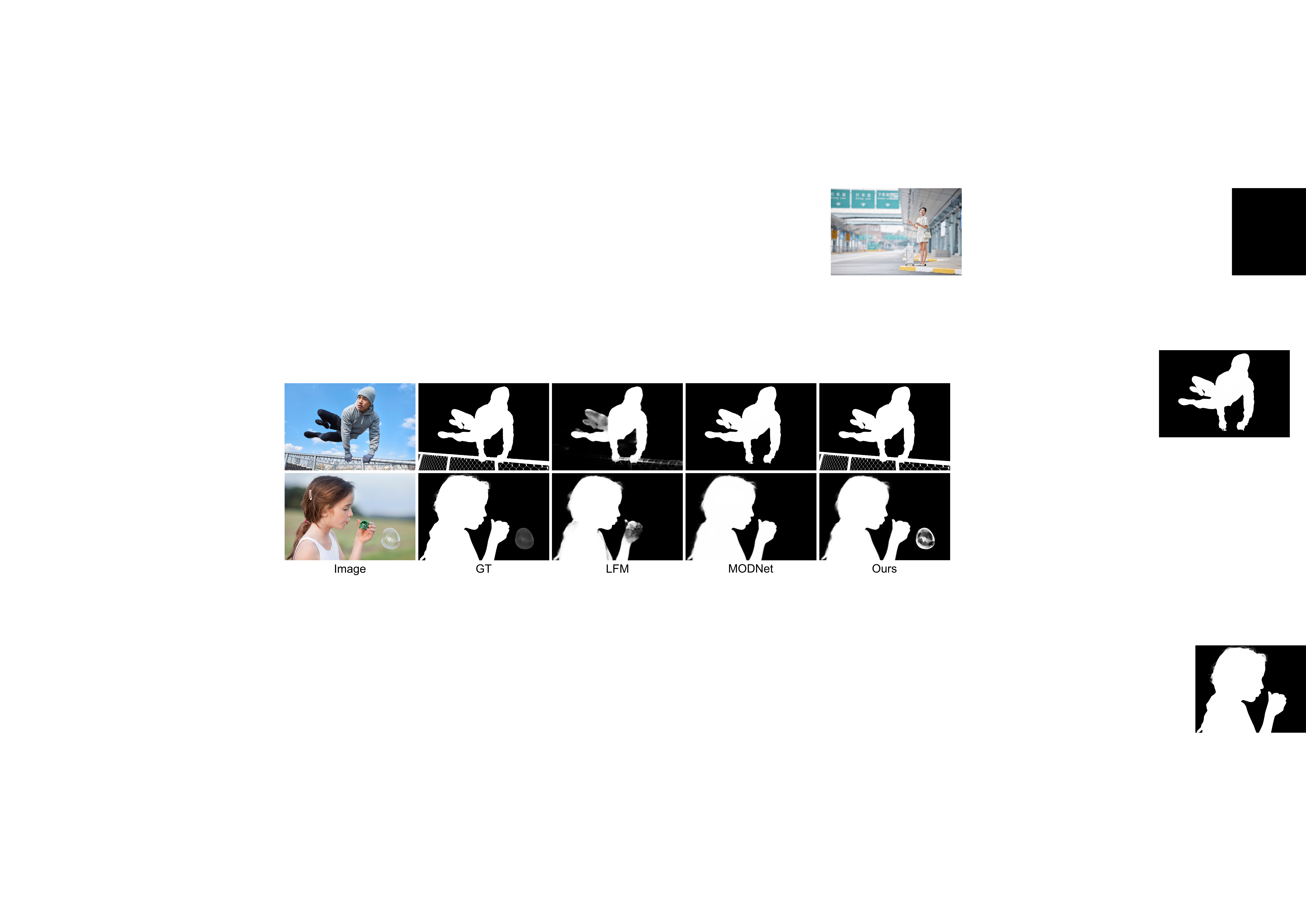}
  \caption{Visual comparisons of our SPG-IM with other trimap-free matting methods (LFM~\cite{zhang2019late} and MODNet~\cite{ke2020green}) on the Real-World images.}
  %\Description{Enjoying the baseball game from the third-base seats. Ichiro Suzuki preparing to bat.}
  \label{fig:teaser}
\end{teaserfigure}

\maketitle

\section{Introduction}

Image matting is a fundamental computer vision task with great application value, which aims to separate the foreground from a single image or video stream and then composites it with a new background. It has been practically applied in the background replacement scenarios of the multimedia such as entertainment image/video creation and special-effect film-making, without green screen backgrounds. Due to the rapid development of the deep neural networks in computer vision, automatic matting becomes increasingly matured~\cite{chen2013knn,levin2007closed,lu2019indices,sengupta2020background,xu2017deep,zhang2019late,Qiao_2020_CVPR,liu2021tripartite,lin2021real}. 

However, there still remain two big challenges in the matting task. First, unlike some well-defined tasks such as object detection or segmentation, image matting is an ill-posed problem that requires a user input trimap or some interactive strokes/points to clearly separate foreground from background, which leads to the inconsistency matting annotation from the annotators' subjective understanding of the foreground. If we present the same image to a group of subjects, it's very natural that each individual will interpret the foreground of this image a bit differently based on his or her educational background, gender, ethnicity, or religious beliefs. Such variation impacts image matting annotations, especially in the human-object interactive and multi-object scenes. For example, as shown in Row 1 of  Figure~\ref{fig:teaser}, annotator A is more willing to interpret the fence as the foreground object together with the man in this image, while annotator B justifiably prefers to only highlight human. One of the most straightforward ways to learn the underlining distribution
of such variation is by introducing large-scale data-driven training. Unfortunately, it is unrealistically expensive to acquire such a representative dataset to cover the real data distribution due to the fine pixel-level granularity annotation of image matting. Recent existing image matting training datasets like AIM~\cite{xu2017deep} and Distinction-646~\cite{Qiao_2020_CVPR} are examples of such data insufficiency. 

Second, for most of the existing automatic matting models without extra inputs ($e.g.$ trimap, interactive strokes/points, or known background), the learning of image saliency steams from object detection or segmentation methodology but lacks the global situational awareness on multiple salient objects and their surrounding environment. For example in Row 2 of Figure~\ref{fig:teaser}, although no touch interaction with the girl, that bubble, which is critical for semantic completeness, is indisputably salient. Unfortunately, previous methods fail to extract such complete and meaningful saliency.

To address the above challenges, in this paper we propose a Situational Perception Guided Image Matting (SPG-IM) network, that aims to mitigate the subjective bias of matting annotations and capture sufficient
situational perception information for global saliency learning. We seek semantic distilled information from the visual-to-textual task to guide the visual features of image matting, due to its large quantity but low-cost training samples. We believe that visual representations from the visual-to-textual task, $e.g.$ image captioning, focus on more semantically comprehensive signals between a)object to object and b)object to the ambient environment to generate descriptions that can cover both the global info and local details. In addition, compared with the expensive pixel annotation of image matting, textual labels can be massively collected at a very low cost.

The SPG-IM network consists of two stages: Situational Perception Distillation (SPD) and Situational Perception Guided Matting (SPGM), both in an encoder-to-decoder fashion. In the first stage, we first pretrain the visual front-end and transformer decoder jointly to generate captions, and aim to learn visual representations including situational perception from visual-to-textual feature transformation. Then the visual front-end is spliced to a new back-end for saliency foreground mask prediction. In the second stage, the SPGM module takes both generated mask and the raw RGB image as inputs and outputs the estimated alpha matte. To leverage situational perception guidance, we propose a textual Semantic Transformation module that transforms and integrates the visual feature stream of the SPD module to guide the visual representations of the SPGM module at multi-scale levels. In addition, we propose an Adaptive Focal Transformation (AFT) Refinement Network that can adaptively select the size of the receptive field to process global context and local details separately, aiming to reach a good balance between complementary global information and local attributes when processing the fused situational perception guided visual features. To justify our solutions, we compare our algorithm, objectively and subjectively, with other methods. Also, we demonstrate by ablation study that the introduction of visual-to-textual transform as semantic guidance can mitigate subjective annotation bias and improve matting performance by leveraging inexpensive image captioning labeling.

Overall, the contributions of this paper are as follows:
\begin{itemize}
    \item To the best of our knowledge, we are the first to underscore the subjective nature of foreground saliency in image/video matting and accordingly introduce situational perception guidance from the visual-to-textual transformation with low-cost labeling to semantically guide the visual features to compensate demographic bias and improve matting performance.
    %\vspace{-6pt}
    \item We build a large-scale matting dataset consisting of 1000 images and corresponding alpha mattes for multi-foreground-object scenes. To the best of our knowledge, this is the first large-scale and high-quality dataset for multi-foreground-object scenes.
    \item We propose a textual Semantic Transformation (TST) module to effectively transform and integrate more situational perception information that guides the matting network.
    %\vspace{-6pt}
    \item We propose an Adaptive Focal Transformation (AFT) Refinement Network to adaptively select the size of receptive fields and focal regions to simultaneously improve global and local performance.
    %\vspace{-6pt}
    \item Extensive experiments demonstrate the effectiveness of our situational perception-guided image matting method, outperforming the state-of-the-art (SOTA) approaches on both synthetic and real-world images.
    %\vspace{-6pt}
\end{itemize}

\section{Related works}

Currently, the matting is generally formulated as an image composite problem, which solves the 7
unknown variables per pixel from only 3 known values:
\begin{equation}
    I_{i} = \alpha_{i} F_{i} + (1-\alpha_{i})B_{i}
\label{E0}
\end{equation}
where 3 dimensional RGB color $I_{i}$ of pixel $i$, while foreground
RGB color $F_{i}$, background RGB color $B_{i}$, and matte estimation $\alpha_{i}$ are unknown. In this section, we discuss the SOTA works trying to solve this under-determined equation. 

\subsection{Classic methods}
Classic foreground matting methods can be generally categorized into two approaches: sampling-based and propagation-based. Sampling-based methods~\cite{aksoy2017designing,chen2013image,shahrian2013improving,wang2007optimized,he2011global,chuang2001bayesian} sample the known foreground and background color pixels, and then extend these samples to achieve matting in other parts. Various sampling-based algorithms are proposed, $e.g.$ Bayesian matting\cite{chuang2001bayesian}, optimized color sampling~\cite{wang2007optimized}, global sampling method~\cite{he2011global}, and comprehensive sampling~\cite{shahrian2013improving}. Propagation-based methods~\cite{levin2008spectral,chen2013knn,he2010fast,lee2011nonlocal,levin2007closed,sun2004poisson} reformulate the composite Eq.~\ref{E0} to propagate the alpha values from the known foreground and background into the unknown region, achieving more reliable matting results.~\cite{wang2008image} provides a very comprehensive review on various matting algorithms. 

\subsection{Deep learning-based methods}
Classic matting methods are carefully designed to solve the composite equation and its variant versions. However, these methods heavily rely on chromatic cues, which leads to bad quality when the color of the foreground and background show small or no noticeable difference.

\begin{figure*}[t]
  \centering
  \includegraphics[width=\linewidth]{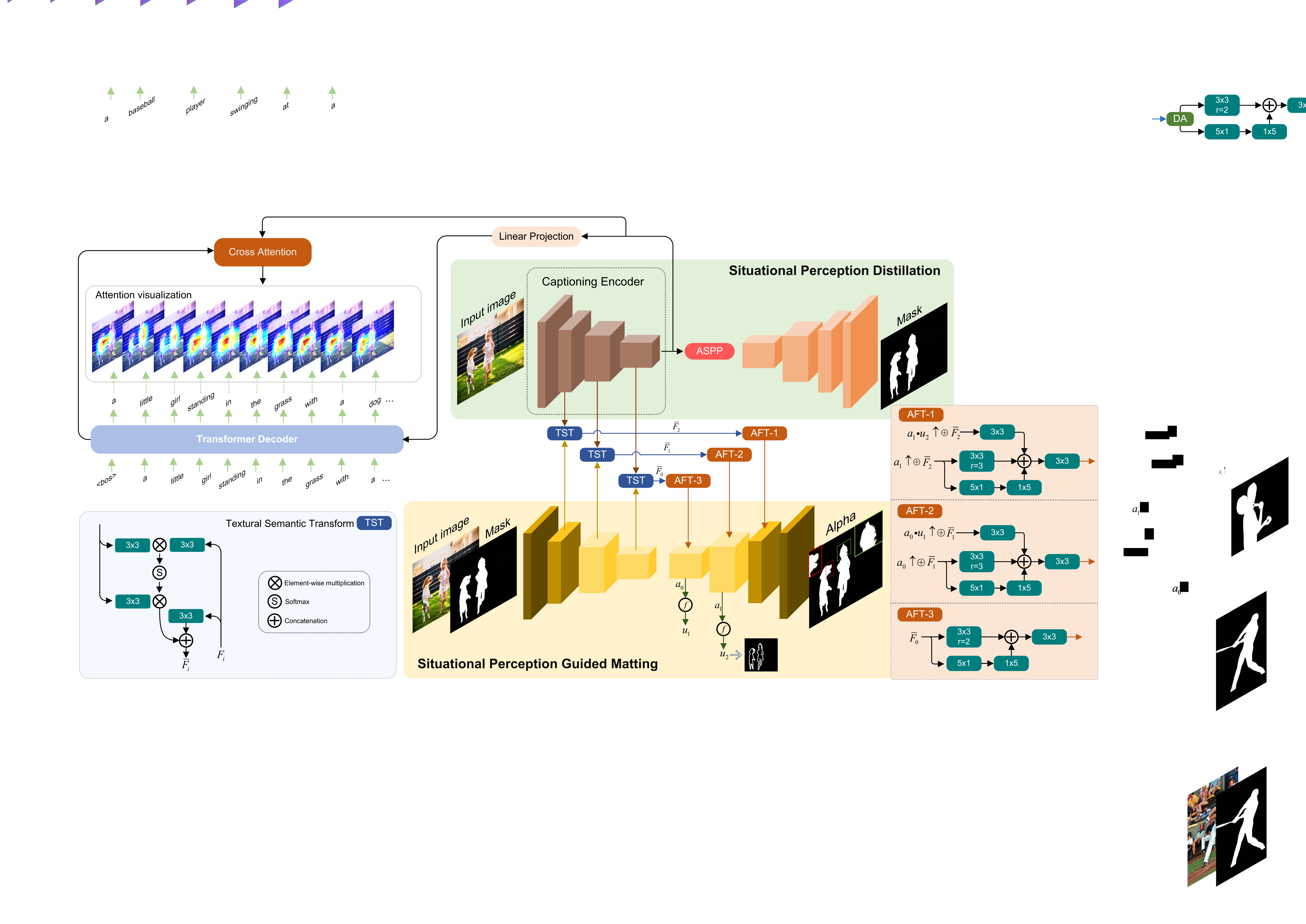}
  \caption{The architecture of the Situational Perception Guidance Image Matting (SPG-IM) network. The SPG-IM network consists of two branches: Situational Perception Distillation (SPD) and Situational Perception Guided Matting (SPGM). The SPD branch takes an RGB image as input and outputs the semantic distilled saliency mask. The SPGM then revisits the raw image input and combines it with the estimated saliency mask for alpha matte prediction, under the guidance of situational perception from the SPD. Both SPD and SPGM employ ResNet-50 as the encoder for visual representation extraction. We utilize ASPP~\cite{chen2017rethinking} to extract and fuse multi-scale contextual information for semantic mask estimation.}\label{fig:pipeline}
  %\Description{A woman and a girl in white dresses sit in an open car.}
\end{figure*}

{\bf Trimap-based methods.} Automatic and intelligent matting algorithms are emerging, due to the rapid development of the deep neural network in computer vision. Initially, some attempts were made to combine deep learning networks with classic matting techniques, $e.g.$ closed-form matting~\cite{levin2007closed} and KNN matting~\cite{chen2013knn}. Cho \etal~\cite{cho2016natural} employ a deep neural network to improve the results of the closed-form matting and KNN matting. These attempts are not end-to-end, so not surprisingly the matting performance is limited by the convolution back-ends. Subsequently, full DL image matting algorithms appear~\cite{chen2018semantic,chen2018tom,xu2017deep}. Xu \etal~\cite{xu2017deep} propose a two-stage deep neural network (Deep Image Matting) based on SegNet~\cite{badrinarayanan2017segnet} for alpha matte estimation and contribute a large-scale image matting dataset (Adobe dataset) with ground truth foreground (alpha) matte, which can be composited over a variety of backgrounds to produce training data. We also use this data for the first-step pre-training of our network. Lutz \etal ~\cite{lutz2018alphagan} introduce a generative adversarial network (GAN) for natural image matting and improve the results of Deep Image Matting~\cite{xu2017deep}. Cai \etal ~\cite{cai2019disentangled} investigate the bottleneck of the previous methods that directly estimate the alpha matte from a coarse trimap, and propose to divide the matting problem into trimap adaptation and alpha estimation tasks. Hou \etal ~\cite{hou2019context} employs two encoder networks to extract essential information for matting, however it is not robust to faulty trimaps. Forte \etal~\cite{lin2022robust} propose a low-cost upgrade to alpha matting networks to also predict the foreground and background colours. They study variations of the training regions and explore a wide range of existing and novel loss functions for optimal prediction. Liu \etal~\cite{liu2021tripartite} propose a 3-branch encoder to accomplish comprehensive mining of the input RGB image and its corresponding trimap, and then develop a Tripartite Formation Integration (TI2) Module to transform and integrate the interconnections between the different branches.

{\bf Additional natural background.} Qian \etal~\cite{qian1999video} compute a probability map to classify each pixel into the foreground or background by simple background
subtraction. This algorithm is sensitive to the threshold and fails when the colors of the foreground and background are similar. Sengupta \etal~\cite{sengupta2020background} introduce a self-supervised adversarial approach - Background Matting (BGM), achieving state-of-the-art results. However, as a prerequisite, the photographer needs to take a shoot of natural background first, which is not friendly to the intensive multi-scene shooting application. Liu~\etal\cite{lin2021real} propose the Background Matting V2 that employs two neural networks: a base network computes a low-resolution result which is refined by a second network operating at high-resolution on selective patches.

{\bf Trimap-free methods.} Currently, a majority of deep image matting algorithms~\cite{cai2019disentangled,hou2019context,lutz2018alphagan,xu2017deep} try to estimate a boundary that divides the foreground and background, with the aid of a user-generated trimap. Several trimap-free matting methods~\cite{chen2018semantic,zhang2019late} predict the trimap first, followed by alpha matting. Qiao \etal~\cite{Qiao_2020_CVPR} employ spatial and channel-wise attention to integrating appearance cues and pyramidal features, they also introduce a hybrid loss function fusing Structural SIMilarity (SSIM), Mean Square Error (MSE), and Adversarial loss to guide the network to further improve the overall foreground structure in trimap-free matting. Lin~\etal~\cite{lin2022robust} propose a robust real-time matting method (RVM) training strategy that optimizes the network on both matting and segmentation tasks. Ke~\etal~\cite{ke2020green} present a lightweight matting objective decomposition network (MODNet) by optimizing a series of sub-objectives simultaneously via explicit constraints. They also introduce an e-ASPP module to fuse multi-scale features, plus a self-supervised sub-objectives consistency (SOC) strategy to address the domain shift problem, which is common in trimap-free methods.

Besides, most current trimap-free methods focus only on human/portrait matting but ignore the objects that are interacting with or attached to people. In addition, they learn the saliency of the images by data-driven training, which lacks the situational perception between salient objects and the surrounding environment, leading to biased or incomplete foreground prediction, especially in multi-object scenes. This is the main reason why we propose the method of Situational Perception Guided Image Matting. In this paper, we quantitatively evaluate the performance of our model for alpha matting in human-object interactive and multi-object scenes.

\subsection{Image Captioning.}
    The problem of generating natural language descriptions from visual data has long been studied in computer vision. 
    Early methods use pre-defined templates, such as object detectors and attribute predictors to generate captions~\cite{socher2010connecting,yao2010i2t}.
    With the rise of Deep Learning-based networks, RNNs~\cite{rennie2017self,vinyals2016show} are adopted as language models to decode corresponding visual features. 
    
    Due to the wide success of transformers in natural language processing and multi-media, image caption methods use transformers to either generate captions directly or fuse visual and language features.
    Herdade~\etal~\cite{herdade2019image} propose a object relation transformer and build image captions based on inter-object relations.
    Liu~\etal~\cite{li2019entangled} introduce an enTangled attention-based transformer that simultaneously exploits visual and semantic information.
    Huang~\etal~\cite{huang2019attention} propose an attention model that first generates an information vector and an attention gate, and then adds secondary attention using element-wise multiplication to aggregate the attended features.
    
    Recent works have demonstrated that image captions can guide feature learning on various visual tasks.
    Karan Desai~\etal~\cite{desai2021virtex} propose a pre-train approach using semantic dense captions to learn visual representations.
    We believe distilled semantic caption information can guide the feature learning of salient object detection tasks.

\section{Methodology}

The network architecture of the Situational Perception Guided Image matting (SPG-IM) is designed to automatically extract the accurate saliency foreground from an RGB image instead of using interactive strokes/points or extra inputs, $e.g.$ trimap and known background. The SPG-IM network consists of two branches: Situational Perception Distillation (SPD) and Situational Perception Guided Matting (SPGM). We pretrain the front-end of SPD under an image caption generation framework before it is transferred to the downstream semantic distilled saliency mask estimation task. The SPGM then revisits the raw image input and combines it with the estimated saliency mask for alpha matte prediction, under the guidance of situational perception from the SPD. The overall architecture of the SPG-IM network is shown in Figure~\ref{fig:pipeline}.

\subsection{Situational Perception Distillation}\label{SPD}

The front-end of our Situational Perception Distillation (SPD) branch is pretrained joint with a transformer-based textual decoder ~\cite{desai2021virtex} to generate the textual description of an image. 

At the visual-to-textual pretraining stage, the Transformer-based textual decoder decodes the visual features output by the visual front-end (captioning encoder) and generates a corresponding image caption $C = (c_0, c_1, ..., c_T, c_{T+1})$. The start of the image caption is $c_0 = [SOS]$,while $c_{T+1} = [EOS]$ indicates the end of a caption sequence.

\textbf{Visual front-end (captioning encoder).} The visual encoder uses a convolutional network to compute the downsampled visual features. For the input image $I$, we use the ResNet-50~\cite{he2016deep} as the visual encoder to extract grid feature $C\in\mathbb{R}^{2048\times(7\times 7)}$, followed by a linear projection layer before sending it to the textual decoder.

\textbf{Textual decoder:} The textual decoder receives a set of grid visual features and outputs the corresponding image caption. We predict the captions in both forward and backward manner and utilize Transformer~\cite{vaswani2017attention} as the backbone of the textual decoder, which adopts a self-attention and cross attention mechanism to fuse visual features using textual queries.

The inputs of the textual decoder module are a set of image features from the visual encoder and a list of caption tokens.
Grid visual features fed into the textual decoder are tokenized to a sequence of the patch features $G\in\mathbb{R}^{D_I\times N_I}$, where each $N_I = 7\times 7$ patch has a feature vector with $D_I-$dimension.   

The first token $c_0 = [SOS]$ indicates the start of the sentence. The transformer backbone iteratively predicts each word in the caption sentence. The prediction ends when transformer output $C_{T+1} = [EOS]$ label. We visualize the word-level attention maps in the cross attention module, where the highlighted regions illustrate that the visual representations from the visual-to-textual task can focus more on the global situational perception. More implementation details are described in the supplementary material.

\textbf{Situational perception distillation.} We transfer the visual features learned by the visual-to-textual transformation to the downstream dense prediction task. We adopt an encoder-to-decoder framework in the SPD network. An Atrous Spatial Pyramid Pooling (ASPP)~\cite{chen2017deeplab} module is set between the visual front-end and decoder to enhance the fusion capabilities of multi-scale features for semantic situational perception. The semantic distilled saliency mask output by the SPD network is supervised by the Gaussian transformed thumbnail of the ground truth alpha matte at a $L_{2}$ loss:
\begin{equation}\label{eq_SPD}
L_{SPD} = \left \|  M_{s}-S(a^{*}) \right \|_{2}
\end{equation}
where $a^{*}$ is the ground truth of alpha matte, $S(a^{*})$ is a Gaussian blur operation after the downsampling of $a^{*}$. We utilize $L_{2}$ loss to smooth the boundary and details of the estimated semantic distilled saliency mask $M_{s}$. Then $M_{s}$ is fed into the SPGM branch along with the raw image for alpha prediction.

\subsection{Situational Perception Guided Matting}
The situational perception guided matting (SPGM) branch receives the RGB image $I$ and the semantic distilled saliency mask $M_{s}$ as inputs to transfer the high-level semantic generalizations into fine-grained foreground alpha mattes. The coarse foreground masks are experimentally proven to be effective as semantic priors for image matting in previous works~\cite{sengupta2020background,yu2021mask}. As for situational perception guidance, we propose a textual Semantic Transformation (TST) module that effectively transforms and integrates the visual feature stream of the SPD module, and guides the visual representations of the SPGM module at multi-scale levels.

\textbf{Textual Semantic Transformation.} The textual Semantic Transformation (TST) module is performed in a non-local fashion, its network architecture is shown in Figure~\ref{fig:pipeline}. We first encode the visual presentation $F_{i}'$ of SPD and the visual representation $F_{i}$ of SPGM separately into the key ($k_{i}'$, $k_{i}$) maps and value ($v_{i}'$, $v_{i}$) maps at each feature scale. The fusion attention map $f_{i}$ is computed by comparing the pixel-by-pixel similarity between $k_{i}'$ and $k_{i}$:
\begin{equation}
    f_{i} = softmax(k{}'k^\top)
\end{equation}
Then we utilize $f_{i}$ as indexes to retrieve the effective situational perception information and concatenate it with $v_{i}$ to update the visual representation of SPGM:     
\begin{equation}
\bar{F_{i}} = v_{i} \oplus(v{}_{i}'^\top \odot f_{i})
\end{equation}
where $\odot$ denotes an element-wise multiplication and $\oplus$ denotes the concatenation. We conduct textual semantic transformation on the features from ResNet50’s ${2,3,4}-th$ res blocks. Consequently, the semantic distilled situational perception information can work as a guiding role at multiple feature levels.

%在日常生活中，当我们用眼睛观察某个目标物体时，我们会用大的感受野去物体的整体，当我们为了更清晰地观察边缘细节时，如毛发和轮廓，我们会动态变换焦点来聚焦到局部细节上   我们发现，对于边界区域需要聚焦在低级别的局部特征特征上

%你这一段到底是想说人会先看整体再看局部，还是说人的观察会在整体和局部之间反复切换？这个是不是改为adaptive focal transformation更贴切？
\textbf{Adaptive Focal Transformation Refinement Network.} As described in previous dense prediction work~\cite{peng2017large}, larger receptive fields establish
dense connections between feature maps and per-pixel classifiers which improve the accuracy of internal regions, while smaller receptive fields benefit the localization focus on local fine-grained details near the object boundaries. It is impossible to find correct and accurate local details if the network is already confused on the global or regional level, which can only generate fine-grained but semantically incorrect results at best. However, it's equally impossible to get good matting if the network is heavily biased to global or regional features since that can only cause image blur.

%这里我重写了一下
That holds also true for the human visual system. When humans observe objects with their eyes, they usually use an adaptive focal strategy to first capture the main body and then narrow their vision on the particular details, such as hair or other small texture. After accumulating a few scrutinized details, the human in return re-gauge or re-evaluate their general perception on those objects. Inspired by this, we propose an Adaptive Focal Transformation (AFT) Refinement Network that can adaptively switch dimensions of receptive fields and focal regions, aiming to complement global information and local attributes when processing the fused situational perception guided features. Specifically, we first generate the focal region mask $u_{i}$ at $ith$ level from the output $a_{i-1}$ of the previous feature level by the following formula:
\begin{equation}
u_{i}(x,y) = \left\{\begin{matrix}
1 & if\ 0 < a_{i-1}(x,y) < 1\\ 
0 & otherwise
\end{matrix}\right. 
\end{equation}
where we set the low confidence regions $0 < a_{i-1}(x,y) < 1$ which consists mainly of boundary details as shown in Figure~\ref{fig:pipeline} and need to be adaptively focused and progressively refined. Then we upsample $a_{i-1}$ and $a_{i-1} \cdot u_{i}$, and concatenate them on the fused situational perception guided feature $\bar{F_{i}}$ respectively at $ith$ level. As shown in the pink box of Figure~\ref{fig:pipeline}, we apply multi-size kernels, larger sizes for $(a_{i-1}\uparrow\oplus\bar{F_{i}})$ to predict main body regions, and smaller sizes for $(a_{i-1} \cdot u_{i}\uparrow\oplus\bar{F_{i}})$ to refine boundary regions. Then additional convolutions are utilized to fuse of main body and refined boundary details. To reduce the computation cost, we apply Atrous convolution~\cite{chen2018encoder} kernels along with spatially separable convolution kernels for large receptive field at multi-scale features.

\textbf{Loss function.} For the supervision of the SPGM branch, we only utilize alpha loss to supervise the outputs of different levels in order to verify the validity of this pattern and to prevent bias caused by other losses:  
\begin{equation}
\mathcal{L} = \sum_{i}^{}\lambda_{i} \left \|  a_{i}-a^{*} \right \|_{1}
\end{equation}
where $\lambda_{i}$ is the loss weight assigning to the output alpha $a_{i}$ of the $ith$ level, $a^{*}$ is the ground truth alpha.

\section{Experiments}
We first describe the datasets used for training and testing. Subsequently, we compare our results with existing state-of-the-art (SOTA) foreground matting algorithms. Finally, we conduct ablation experiments to show the effectiveness of each branch and module. More implementation details are provided in the supplementary material.

\subsection{Datasets}
\textbf{Composition-1K~\cite{xu2017deep}.} The training set consists of 431 foreground objects and each of them is composited over 100 random COCO~\cite{lin2014microsoft} images to produce 43.1k composited training images. For the test set, we combine each foreground of Composition-1K with 20 random VOC~\cite{everingham2010pascal} images to produce 1k composited testing images. 
%a slight difference from previous methods is that we combine each foreground of Composition-1K with 20 random HDR images from the real world and follow~\cite{lagunas2021single} to relight the foreground to produce realistic composite images. Relighting the foreground helps to verify the ability of matting models to learn a true understanding of the semantic information, rather than possibly just learning the color or tone difference between the foreground and the new background. 
And then we follow Eq.~\ref{eq_SPD} in Section~\ref{SPD} to generate $S(a^{*})$ for supervising the training of the Situational Perception Distillation (SPD) branch.

\textbf{Distinction-646~\cite{Qiao_2020_CVPR}.} It includes 431 and 50 foreground objects in training and test sets, respectively. We enforce the same rule and composited ratio with the Composition-1K. 

\textbf{Human-2K~\cite{liu2021tripartite}.} It provides 2100 foreground images (2000 for training and 100 for testing). The same rules and ratios as Composition-1K are used in the Human-2K to composite new images.

\begin{figure*}[t]
  \centering
  \includegraphics[width=1.0\linewidth]{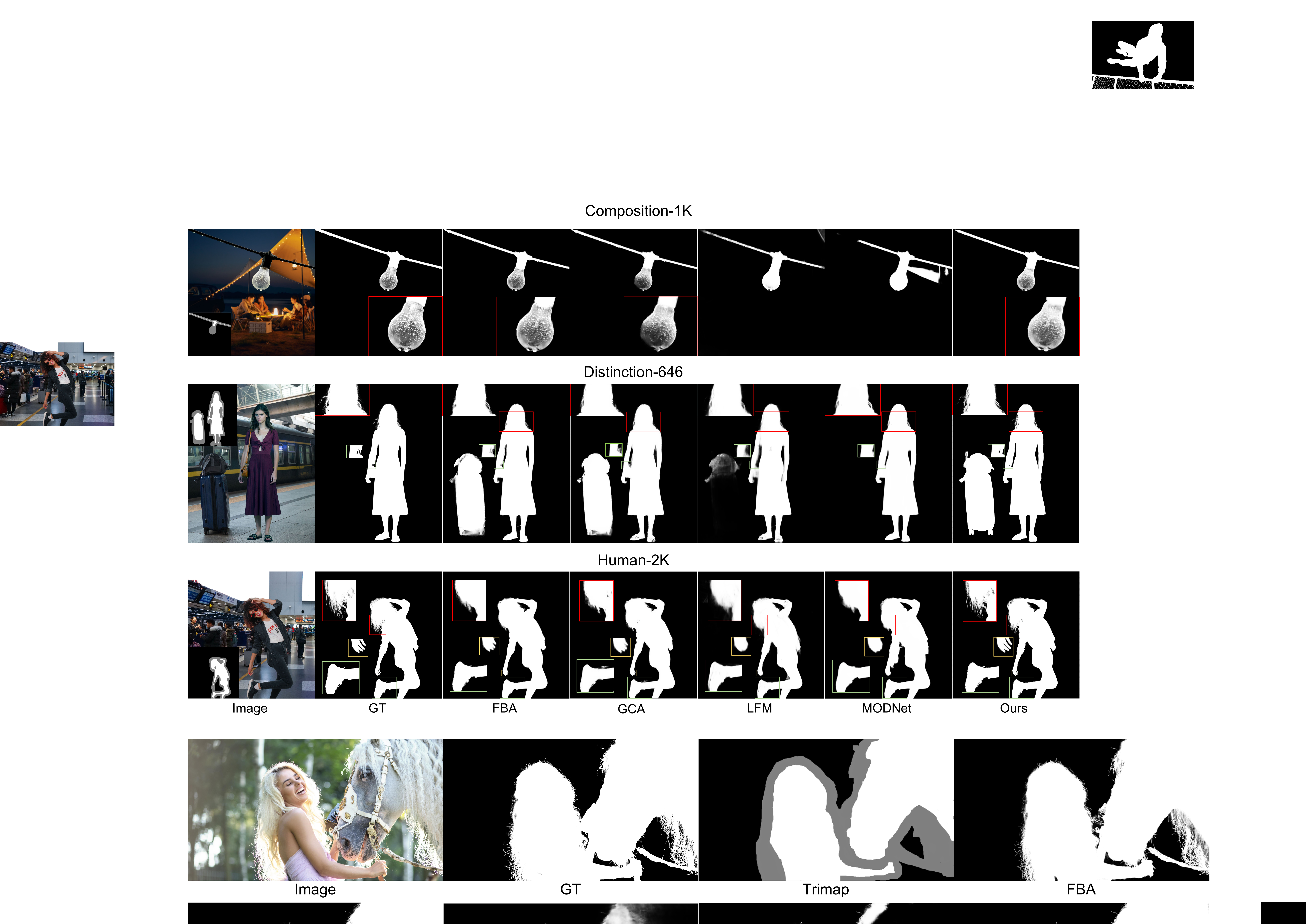}
  \caption{Visual comparison on public composition datasets. Trimap-based methods: FBA~\cite{forte2020f} and GCA~\cite{li2020natural}. Trimap-free methods: LFM~\cite{zhang2019late}, MODNet~\cite{forte2020f}, and ours.}\label{fig:visual_public}
  %\Description{A woman and a girl in white dresses sit in an open car.}
\end{figure*}

\begin{table}[t]
\renewcommand\tabcolsep{8.0pt}
\begin{center}
\begin{tabular}{lcccc}
\toprule
{\bf Methods}&SAD$\downarrow$& MSE$\downarrow$& Grad$\downarrow$& Conn$\downarrow$\\
\midrule
DIM~\cite{xu2017deep}&50.4& 0.014&31.0&50.8 \\
IndexNet~\cite{lu2019indices}&45.8& 0.013&25.9&43.7 \\
SampleNet~\cite{tang2019learning}&40.4&0.010&-&-\\
CAM~\cite{hou2019context}&35.8& 0.008&17.3&33.2 \\
LFM~\cite{zhang2019late}&49.0& 0.020&34.3&50.6 \\
HAttMatting~\cite{Qiao_2020_CVPR}&44.0& 0.007&29.3&46.4 \\
%BSHM~\cite{liu2020boosting}&13.92& 4.67&70.32&10.02 \\
MODNet~\cite{forte2020f}&43.7& 0.012&29.1&42.0 \\
GCA~\cite{li2020natural}&35.3& 0.009&16.9&32.5 \\
%GFM~\cite{li2022bridging}\\
HDMatt~\cite{yu2021high}&33.5&0.007&14.5&29.9\\
SIM~\cite{sun2021semantic}&28.0&0.006&10.8&24.8\\
FBA~\cite{forte2020f}&25.8&0.005&10.6&20.8\\
TIMI-Net~\cite{liu2021tripartite}&29.1&0.006&11.5&25.4\\
    %\cmidrule(l){2-6}
MG-$Mask_{SPD}$~\cite{yu2021mask}&30.2&0.007&12.9&26.6\\
%BGMV2~\cite{lin2021real}&14.61&4.92&75.68&10.61\\
MODNet~\cite{forte2020f}&43.7& 0.012&29.1&42.0 \\
    %\cmidrule(l){2-6}
\midrule
    %\cmidrule(l){2-6}
%{\bf BM}~\cite{chuang2001bayesian}&$T$, $B$& 8.26& 0.0171\\
Ours &{\bf 23.2}&{\bf 0.004}&{\bf 8.9}&{\bf 18.6}\\
\bottomrule
\end{tabular}
\end{center}
%\vspace{-10pt}
\caption{The quantitative results on Composition-1K. $Mask_{SPD}$ denotes the guidance input for MG matting~\cite{yu2021mask}.}
\label{Results_4data}
%\vspace{-10pt}
\end{table}
%\vspace{-10pt}
\begin{table}[t]
\renewcommand\tabcolsep{7.8pt}
\begin{center}
\begin{tabular}{lcccc}
\toprule
{\bf Methods}&SAD$\downarrow$& MSE$\downarrow$& Grad$\downarrow$& Conn$\downarrow$\\
\midrule
DIM~\cite{xu2017deep}&47.6&0.009&43.3&55.9\\
HAttMatting~\cite{liu2021tripartite}&49.0&0.009&41.6&49.9\\
    %\cmidrule(l){2-6}
TIMI-Net~\cite{liu2021tripartite}&22.3&0.011&14.4&20.5\\
MODNet&46.2&0.009&39.0&43.5\\
MG-$Mask_{SPD}$~\cite{yu2021mask}&23.6&0.007&16.1&21.0\\
%MG&40.35& 16.80 & 125.77 & 33.65\\
    %\cmidrule(l){2-6}
\midrule
    %\cmidrule(l){2-6}
%{\bf BM}~\cite{chuang2001bayesian}&$T$, $B$& 8.26& 0.0171\\
Ours &{\bf 20.9}&{\bf 0.006}&{\bf 11.2}&{\bf 19.8}\\
\bottomrule
\end{tabular}
\end{center}
%\vspace{-10pt}
\caption{The quantitative results on Distinction-646~\cite{Qiao_2020_CVPR}.}
\label{Results_4data_1}
%\vspace{-10pt}
\end{table}
%\vspace{-10pt}
\begin{table}[t]
\renewcommand\tabcolsep{7.8pt}
\begin{center}
\begin{tabular}{lcccc}
\toprule
{\bf Methods}&SAD$\downarrow$& MSE$\downarrow$& Grad$\downarrow$& Conn$\downarrow$\\
\midrule
DIM~\cite{xu2017deep}&7.5& 0.008&6.4&6.7\\
TIMI-Net~\cite{liu2021tripartite}&4.2&0.003&2.1&3.0\\
    %\cmidrule(l){2-6}
MODNet&7.8& 0.008&7.2&7.4 \\
MG-$Mask_{SPD}$~\cite{yu2021mask}&4.4& 0.004 & 2.5 & 3.2\\
    %\cmidrule(l){2-6}
\midrule
    %\cmidrule(l){2-6}
%{\bf BM}~\cite{chuang2001bayesian}&$T$, $B$& 8.26& 0.0171\\
Ours &{\bf 4.0}&{\bf 0.002}&{\bf 2.0}&{\bf 2.8}\\
\bottomrule
\end{tabular}
\end{center}
%\vspace{-10pt}
\caption{The quantitative results on Human-2K~\cite{liu2021tripartite}.}
\label{Results_4data_2}
\end{table}
%\vspace{-10pt}

\begin{table*}[t]
\renewcommand\tabcolsep{8.6pt}
\begin{center}
\begin{tabular}{lcccccccc}
\toprule
\multirow{2}{*}{Methods} & \multicolumn{4}{c}{Single-object} & \multicolumn{4}{c}{Multi-object} \\
        \cmidrule(l){2-5} \cmidrule(l){6-9}
&SAD$\downarrow$& MSE$\downarrow$& Grad$\downarrow$& Conn$\downarrow$&SAD$\downarrow$& MSE$\downarrow$& Grad$\downarrow$& Conn$\downarrow$\\
\midrule
TIMI-Net~\cite{liu2021tripartite}&26.4& 0.012&15.7&30.0 &28.3&0.013&26.1&42.4\\
    %\cmidrule(l){2-6}
MODNet&51.5& 0.014&36.1&56.9&69.2&0.024&60.8&81.3 \\
MG-$Mask_{SPD}$&32.9& 0.017 & 26.4 &47.0&33.5&0.019&32.7&55.6\\

    %\cmidrule(l){2-6}
\midrule
Basic&40.2&0.023&43.2&64.5&47.2&0.024&45.0&68.1\\
Basic+PRN$_{MG}$~\cite{yu2021mask} &29.9&0.015&20.4&41.8&32.4&0.017&31.2&53.5\\
Basic+AFT&26.9&0.012&16.5&30.4&27.3&0.015&19.8&32.9\\
SPG-IM w/o AFT&33.7&0.018&30.6&50.9&26.7&0.014&16.0&32.3\\
SPG-IM &{\bf 22.6}&{\bf 0.008}&{\bf 12.5}&{\bf 24.9}&{\bf 22.7}&{\bf 0.008}&{\bf 13.1}&{\bf 27.7}\\
\bottomrule
\end{tabular}
\end{center}
%\vspace{-10pt}
\caption{The quantitative results on our Multi-object-1K.}
\label{ablation}
%\vspace{-12pt}
\end{table*}

\textbf{Multi-Object-1K.} Although there are several typical datasets we can use for the matting task, most of them include only single-object foregrounds. To extend the image matting to the application of multi-foreground-object scenes, we propose our Multi-Object 1K which consists of both single-foreground-object and multi-foreground-object images. Consequently, this dataset can better evaluate the semantic situational perception ability of our method, especially in multi-object scenes. 

Multi-Object-1K provides 1000+200 real-world images and high accuracy alpha mattes, where 70\% of the datasets are multi-object scenes. We believe Multi-Object 1K can serve as a new challenging benchmark in the image matting area. We also apply the same rules and ratios as Composition-1K on our Multi-Object-1K for data composition.  

%\subsection{Implementation Details.}
\subsection{Comparative study on composition datasets}
We conduct comparative study on three composition benchmarks: Composition-1K~\cite{xu2017deep}, Distinction-646~\cite{Qiao_2020_CVPR}, and  Human-2K~\cite{liu2021tripartite} datasets. We report mean square error (MSE), sum of the absolute difference (SAD), spatial-gradient (Grad), and connectivity (Conn) between predicted and ground truth alpha mattes. Lower values of these metrics indicate better estimated alpha matte. 

We compare our method with state-of-the-art (SOTA) trimap-based methods: DIM~\cite{xu2017deep}, IndexNet~\cite{lu2019indices}, CAM~\cite{hou2019context}, GCA~\cite{li2020natural}, FBA~\cite{forte2020f}, and SIM~\cite{sun2021semantic}; mask-based method: MG Matting~\cite{yu2021mask}; trimap-free methods: LFM~\cite{zhang2019late}, HAttMatting~\cite{Qiao_2020_CVPR}, and MODNet~\cite{forte2020f}. To fairly compare them for fully automatic matting, we utilize our Situational Perception Distillation (SPD) branch to produce the semantic distilled foreground mask $Mask_{SPD}$ for MG matting. For trimap-based methods, we generate trimaps from the ground truth alpha mattes by thresholding and random dilation as discussed in~\cite{xu2017deep}. For methods without publicly available codes, we follow their papers to reproduce the results with due diligence.

Table~\ref{Results_4data} to~\ref{ablation} show the quantitative results of our SPG-IM with other SOTA models on four datasets. Our SPG-IM outperforms all competing trimap-free methods (LFM~\cite{zhang2019late}, HAttMatting~\cite{Qiao_2020_CVPR}, and MODNet~\cite{forte2020f}) by a large margin. Meanwhile, our model also shows remarkable superiority over the state-of-the-art (SOTA) trimap-based and mask-guided methods in terms of all four metrics across the public datasets ($i.e.$ Composition-1K, Distinction-646, and Human-2K), and our Multi-Object-1K benchmark. 

\begin{figure*}[t]
  \centering
  \includegraphics[width=1.0\linewidth]{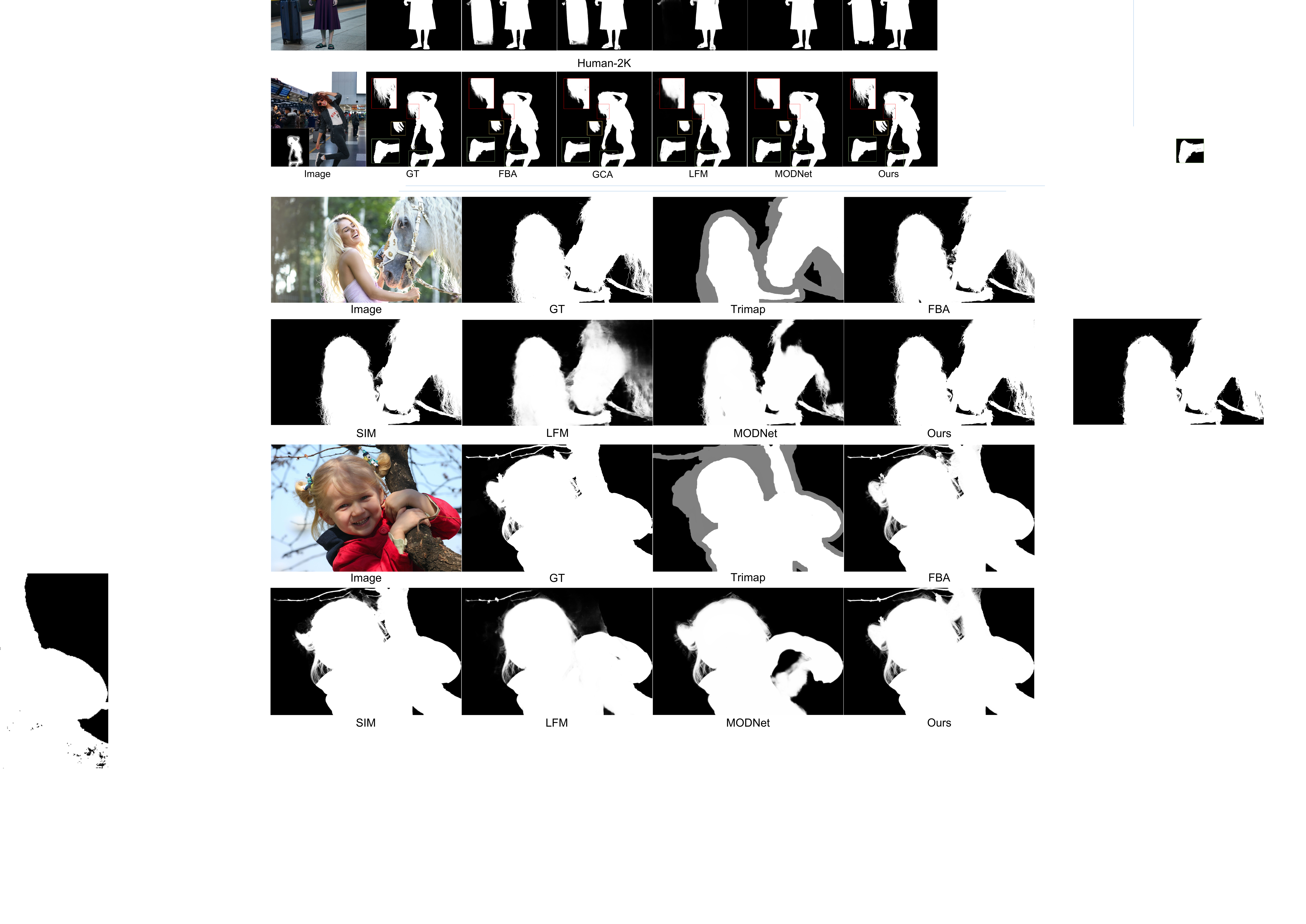}
  \caption{Visual comparisons on our Multi-Object-1K. Trimap-based methods: FBA~\cite{forte2020f} and SIM~\cite{sun2021semantic}. Trimap-free methods: LFM~\cite{zhang2019late}, MODNet~\cite{forte2020f}, and ours}\label{fig:visual_dist_646}
  %\Description{A woman and a girl in white dresses sit in an open car.}
  \label{fig:visual_multi}
%\vspace{-6pt}
\end{figure*}

To give an intuitive understanding of the significance
of the situational perception guidance, we visualize sampled results from our SPG-IM and other SOTA models in Figure~\ref{fig:visual_public} and~\ref{fig:visual_multi}. It can be obviously observed that our method preserves fine details ($e.g.$ hair tip sites, transparent textures, and boundaries) without the guidance of trimap. Moreover, compared to other competing trimap-free models, our SPG-IM can retain better global semantic completeness. For example in Row 2 of Figure~\ref{fig:visual_public}, we composite a human foreground from Distinction-646 into a train station scene with luggage. Due to not directly touching the person, the saliency of the luggage is challenging to capture. Fortunately, with the help of the situational perception guidance, our SPG-IM explores the semantic association between object-to-object ($i.e.$ the person and the luggage) and object-to-environment ($i.e.$ the luggage and the station), and then effectively identifies the saliency of the luggage and considers it as part of the foreground, where other trimap-free methods fail.

\begin{figure}[t]
%\begin{center}
\centering
%\fbox{\rule{0pt}{2in} \rule{.9\linewidth}{0pt}}
    %\qquad\qquad
    \includegraphics[width=1.0\linewidth]{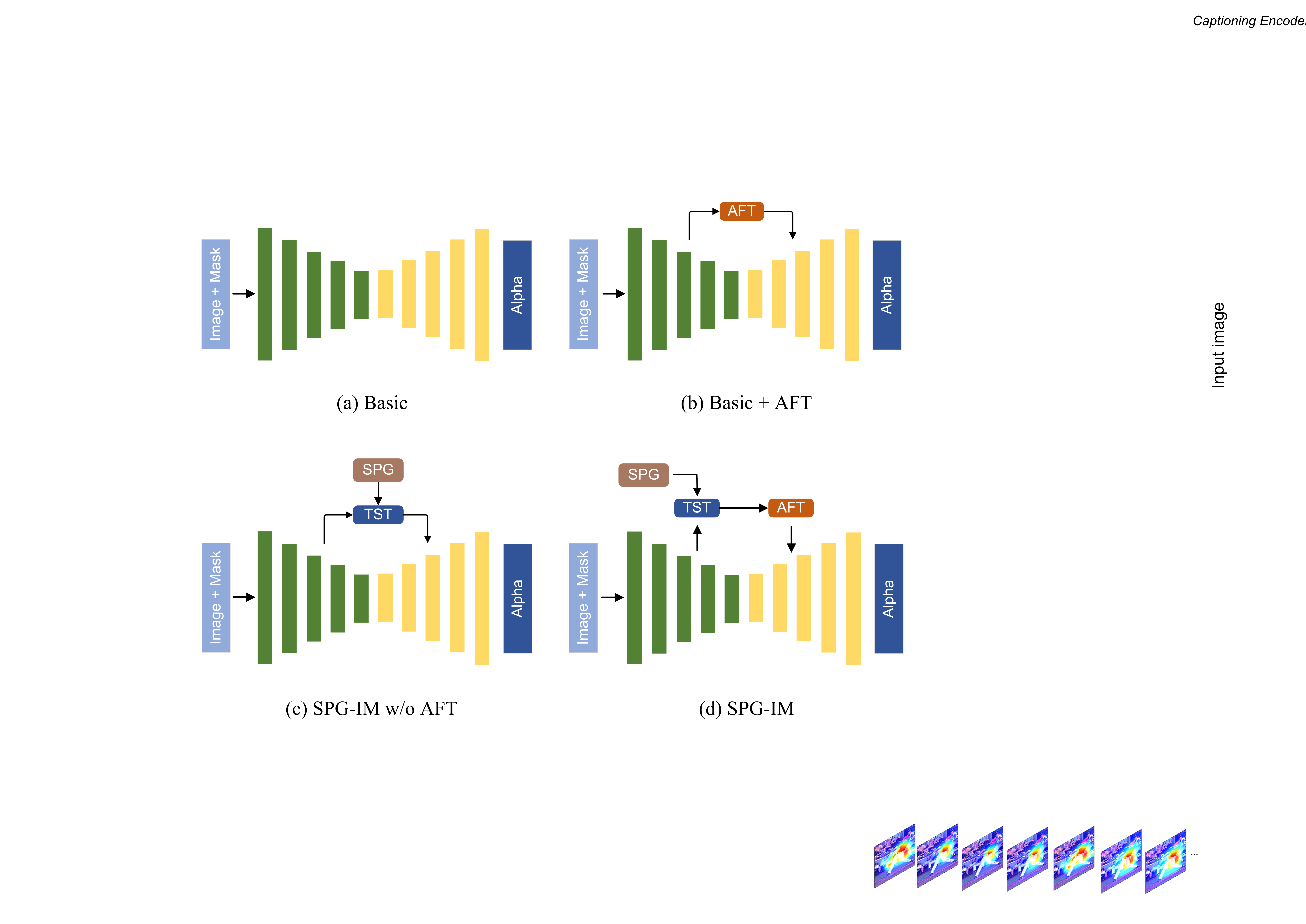}
            %\caption{Caption 1}\label{fig:P3D-a}
\caption{The settings of ablation study for our method.} %We first factor each input image into three virtual modalities, i.e.\ estimated depth map, interaction heatmap and segmentation mask. Then a dual network is employed to predict the same alpha matte with different inputting modalities, followed by a complementary learning module to achieve self supervision across modalities.
\label{fig:ablation_study}
%\vspace{-6pt}
\end{figure}

\subsection{Ablation Study}
We validate the effectiveness of our key components on the Multi-Object-1K dataset, where more than 70\% samples are multi-object foregrounds. As summarized in Table~\ref{ablation}, we conduct quantitative comparisons in both single-object foreground and multi-object foreground scenes respectively. %which consists of large-scale multi-object foregrounds that require more situational perception from object-to-object and object-to-environment to capture complete semantic information.  
The model settings of the ablation study are illustrated in Figure~\ref{fig:ablation_study}, where Basic denotes the encoder-to-decoder structure of the situational perception guidance matting (SPGM) branch without situational perception guidance or Adaptive Focal Transformation (AFT) refinement module. Basic + AFT refers to performing adaptive focal transformation directly on multi-scale features of the Basic network. SPG-IM w/o AFT indicates that the fused situational perception guided feature $\bar{F_{i}}$ is skip-connected respectively to the decoder of SPGM at $ith$ feature level without adaptive focal transformation. 

\textbf{Adaptive Focal Transformation.} We report the quantitative comparison results of our model with and without the Adaptive Focal Transformation (AFT) refinement module at dual settings, $i.e.$ Basic \textbf{\emph{vs.}} Basic+AFT and SPG-IM w/o AFT \textbf{\emph{vs.}} SPG-IM.  As shown in 
Table~\ref{ablation}, both the baseline and its situational perception guided variants show performance gains by applying our AFT module, proving the necessity of the adaptive transformation of receptive fields for focal regions in image matting. We also apply the Progressive Refinement Module (PRM$_{MG}$) from the MG matting~\cite{yu2021mask} on our Basic network for comparison. The quantitative results in Table~\ref{ablation} illustrate the superiority of our AFT which can better complement the high-level semantic information completeness and the low-level subtle detail refinement.

\textbf{Textual Semantic Transformation.} We also evaluate our model under two ablation settings, $i.e.$ Basic \textbf{\emph{vs.}} SPG-IM w/o AFT and Basic + AFT \textbf{\emph{vs.}} SPG-IM. The quantitative comparisons are shown in Table~\ref{ablation}. The proposed TST module improves the performance of alpha estimation in both single-object and multi-object foreground scenes, particularly the improvement in the latter is expectedly significant. This is because the TST module can effectively transform and integrate more situational perception information to guide the matting network for better saliency association between inter-objects and object-to-environment. Additionally, we
demonstrate the performance gain after combining textual semantic transformation and adaptive focal transformation. Some representative visualized comparisons of the original real-world samples from our Multi-Object-1K are provided in Figure~\ref{fig:visual_multi}, which also illustrate that our SPG-IM can enhance both global semantic awareness and local details.  
\vspace{-6pt}
\section{Conclusion}
%\vspace{8pt}
In this paper, we present a situational perception guided matting technique (SPG-IM) that can capture situational perception information distilled from the visual-to-textual task for better global saliency, aiming to compensate the subjective nature of the image matting and mitigate the subjective bias of matting annotations without the expensive pixel annotation.

For the implementation of situational perception guidance, we introduce a Textual Semantic Transformation (TST) module to effectively transform and integrate more situational perception information that guides the matting network. Further, an Adaptive Focal Transformation (AFT) Refinement Network is proposed to adaptively select the size of receptive fields and focal regions to simultaneously improve global and local matting performance. Extensive experiments demonstrate that our model outperforms current state-of-the-art algorithms in both single-object and multi-object foreground scenes without extra inputs, $e.g.$ trimap, known background, and interactive strokes/points. For future works, we will continue to focus on the study of multi-object foreground matting and conduct research on the real-time versions.
% most of the previous methods learn the saliency from object detection or segmentation methodology but lacks the situational global situational awareness on multiple salient objects and their surrounding environment.

\appendix
  \renewcommand{\appendixname}{Appendix~\Alph{section}}
  
\begin{table}[t]
\renewcommand\tabcolsep{3.5pt}
\begin{center}
\scalebox{1.0}{
\begin{tabular}{lr}
\toprule
Parameter & Value  \\
\midrule
%\cmidrule(l){1-1}
Optimizer \qquad\qquad\qquad\qquad\qquad\qquad\qquad & Adam \\
\midrule
\multicolumn{2}{c}{\textbf{SPD}}\\
\cdashline{1-2}[0.8pt/2pt]
Initial learning rate & $1.0 \times 10^{-2}$\\
Input image size & $512 \times 512$\\
Batch size of SPD & 16\\
%Number of SPD epochs & 25\\
\midrule
\multicolumn{2}{c}{\textbf{SPGM}}\\
\cdashline{1-2}[0.8pt/2pt]
Initial learning rate & $5.0 \times 10^{-3}$\\
Input image size & $512 \times 512$\\
Batch Size of SPGM & 4\\
%Number of SPGM epochs & 25\\
Loss weight $\lambda_{1}$ & 1\\
Loss weight $\lambda_{2}$ & 2\\
Loss weight $\lambda_{3}$ & 3\\
\bottomrule
\end{tabular}}
\end{center}
%\vspace{-10pt}
\caption{Implementation details and hyper-parameter setting.}
\label{implementation_details}
%\vspace{-18pt}
\end{table}

\section{Implementation Details}
\subsection{Implementation of the SPG-IM}
The implementation of the Situational Perception Guided Image Matting (SPG-IM) framework is based on the public PyTorch~\cite{paszke2019pytorch} toolbox and trained on a single Tesla V100 GPU with 32GB memory. The training details and all hyper parameters are outlined in Table~\ref{implementation_details}. The learning rate of SPD is decreased by a factor of 10 at the epoch of \{20, 40\}, \{30, 60\}, and \{60, 80\}, and \{40, 60, 80\} for Composition-1K, Distinction-646, and Human-2K, and our Multi-Object-1K, respectively. Meanwhile, the learning rate of SPGM decays at a rate of 0.1 in the epoch of \{20, 30, 40\}, \{40, 60, 80\}, and \{80, 100, 120\}, and \{60, 80, 100\} for Composition-1K, Distinction-646, and Human-2K, and our Multi-Object-1K, respectively.

\subsection{Pretraining of the visual front-end in the SPD branch.}

At the pretraining stage, we train the visual-to-textual network on the COCO Captions dataset~\cite{chen2015microsoft} and use the SGD optimizer with momentum 0.9 and weight decay $10^{-4}$. Follow~\cite{desai2021virtex}, we utilize warmup~\cite{goyal2019scaling} at the beginning iterations followed by cosine decay~\cite{loshchilov2016sgdr} to zero. The max learning rates for the visual front-end and textual decoder are set to 0.2 and $10^{-3}$ respectively. 

\subsection{Model size comparison}
We compare our model size with other trimap-based ($e.g.$ GCA~\cite{li2020natural} and FBA~\cite{forte2020f}), trimap-free ($e.g.$ LFM~\cite{zhang2019late}), and mask-guided ($e.g.$ MG~\cite{yu2021mask}) methods. As shown in Figure~\ref{model_size}, the total model size of our method is smaller than GCA$_{auto}$, FBA$_{auto}$, and MG$_{auto}$. The SPD branch contains 65.7\% of the parameters in our SPG-IM. Similarly, we observe that the automatic generation of higher-level semantic priors (such as trimap and mask) tends to be more computationally intensive for both trimap-based and mask-guided methods. 
For future work, we will optimize the front-end semantic distillation module to achieve the lightweight of the entire model. 

\section{More Visualization Results}
\begin{figure*}[t]
  \centering
  \includegraphics[width=0.91\linewidth]{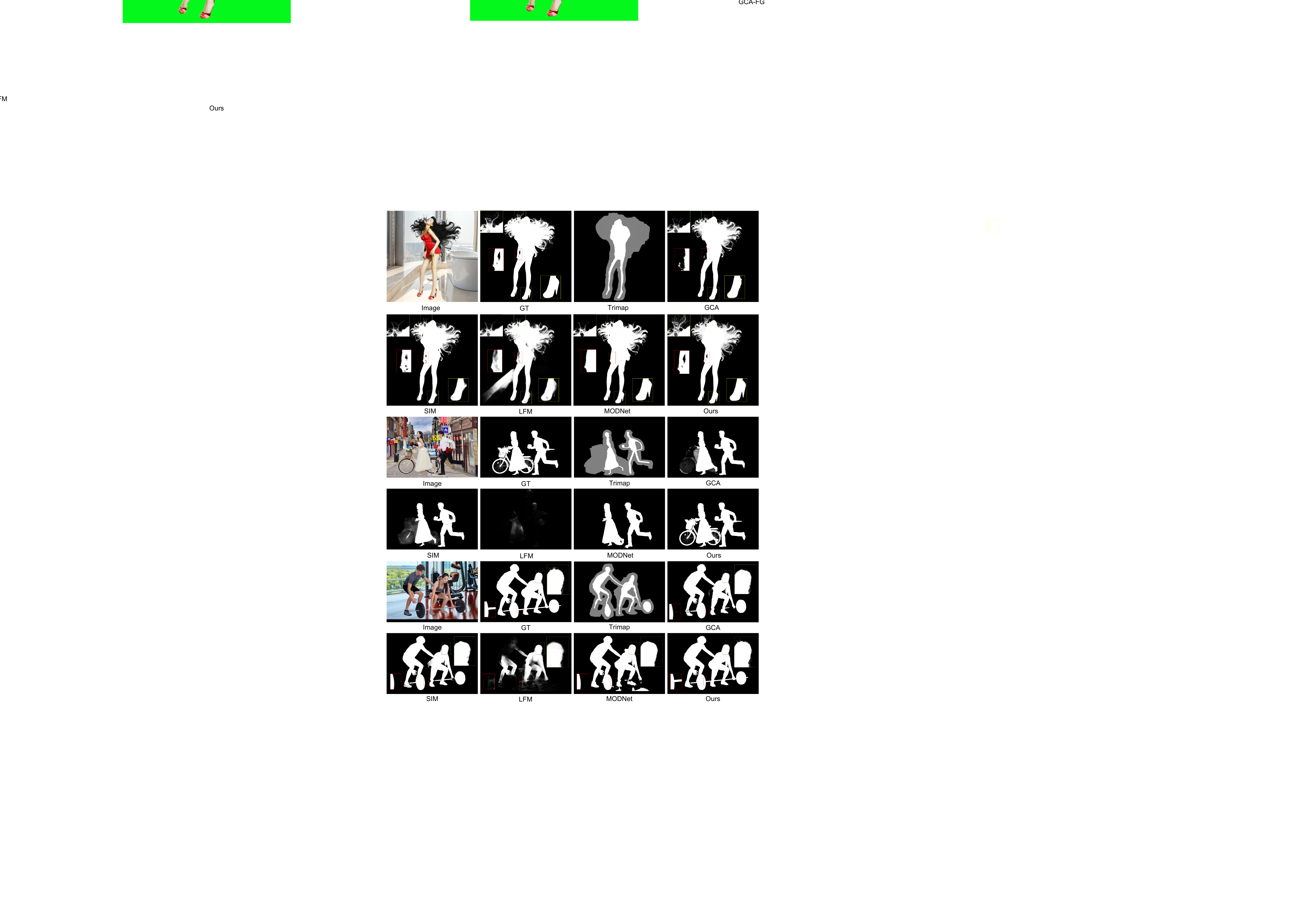}
  \caption{More visual comparisons on our Multi-Object-1K. Trimap-based methods: GCA~\cite{li2020natural} and SIM~\cite{sun2021semantic}. Trimap-free methods: LFM~\cite{zhang2019late}, MODNet~\cite{forte2020f}, and ours.}\label{fig:visual_dist_646}
  %\Description{A woman and a girl in white dresses sit in an open car.}
  \label{fig:visual_multi_1}
%\vspace{-6pt}
\end{figure*}

\begin{figure*}[t]
  \centering
  \includegraphics[width=0.91\linewidth]{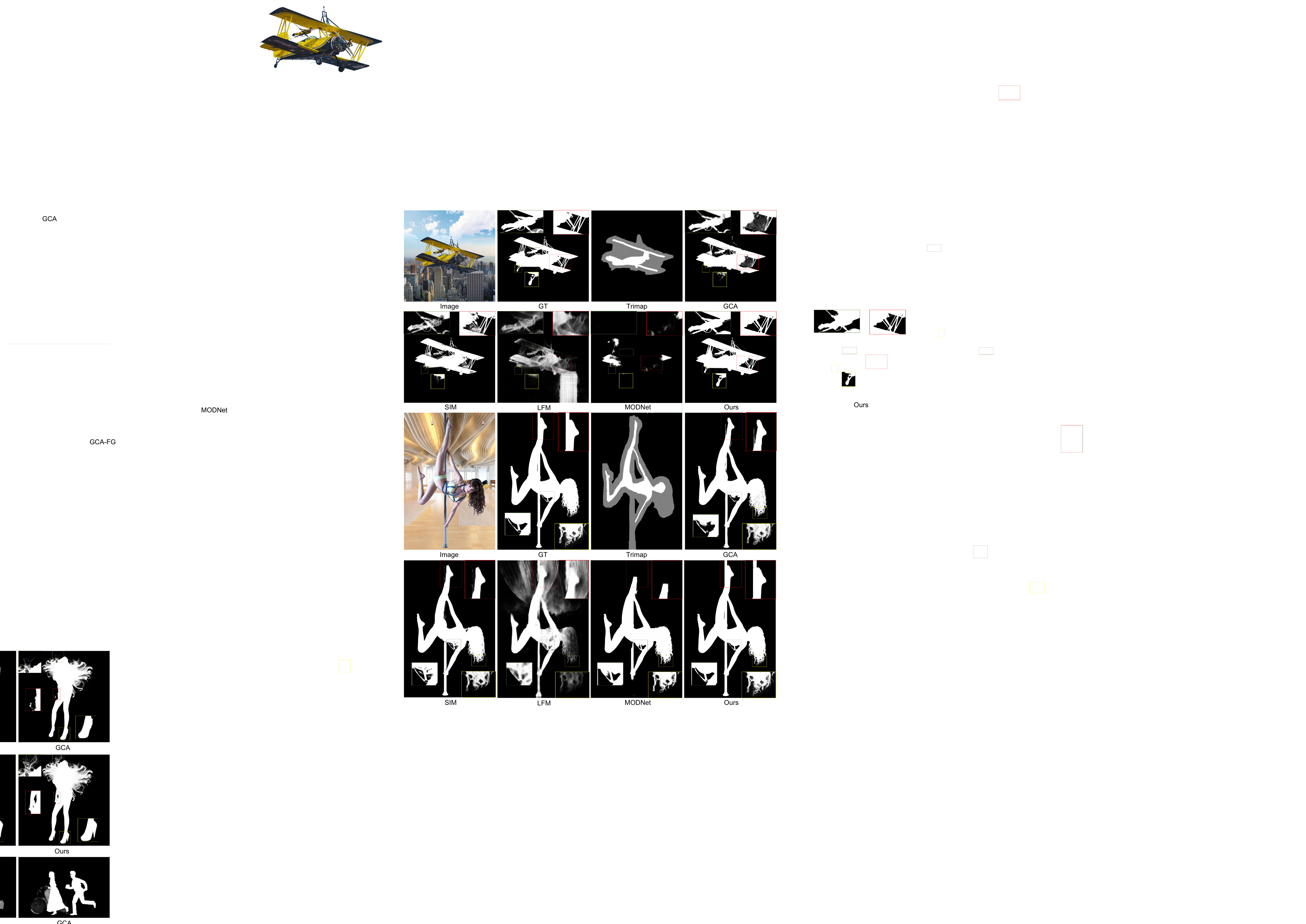}
  \caption{More visual comparisons on our Multi-Object-1K. Trimap-based methods: GCA~\cite{li2020natural} and SIM~\cite{sun2021semantic}. Trimap-free methods: LFM~\cite{zhang2019late}, MODNet~\cite{forte2020f}, and ours.}\label{fig:visual_dist_646}
  %\Description{A woman and a girl in white dresses sit in an open car.}
  \label{fig:visual_multi_2}
%\vspace{-10pt}
\end{figure*}

\begin{figure*}[t]
  \centering
  \includegraphics[width=0.76\linewidth]{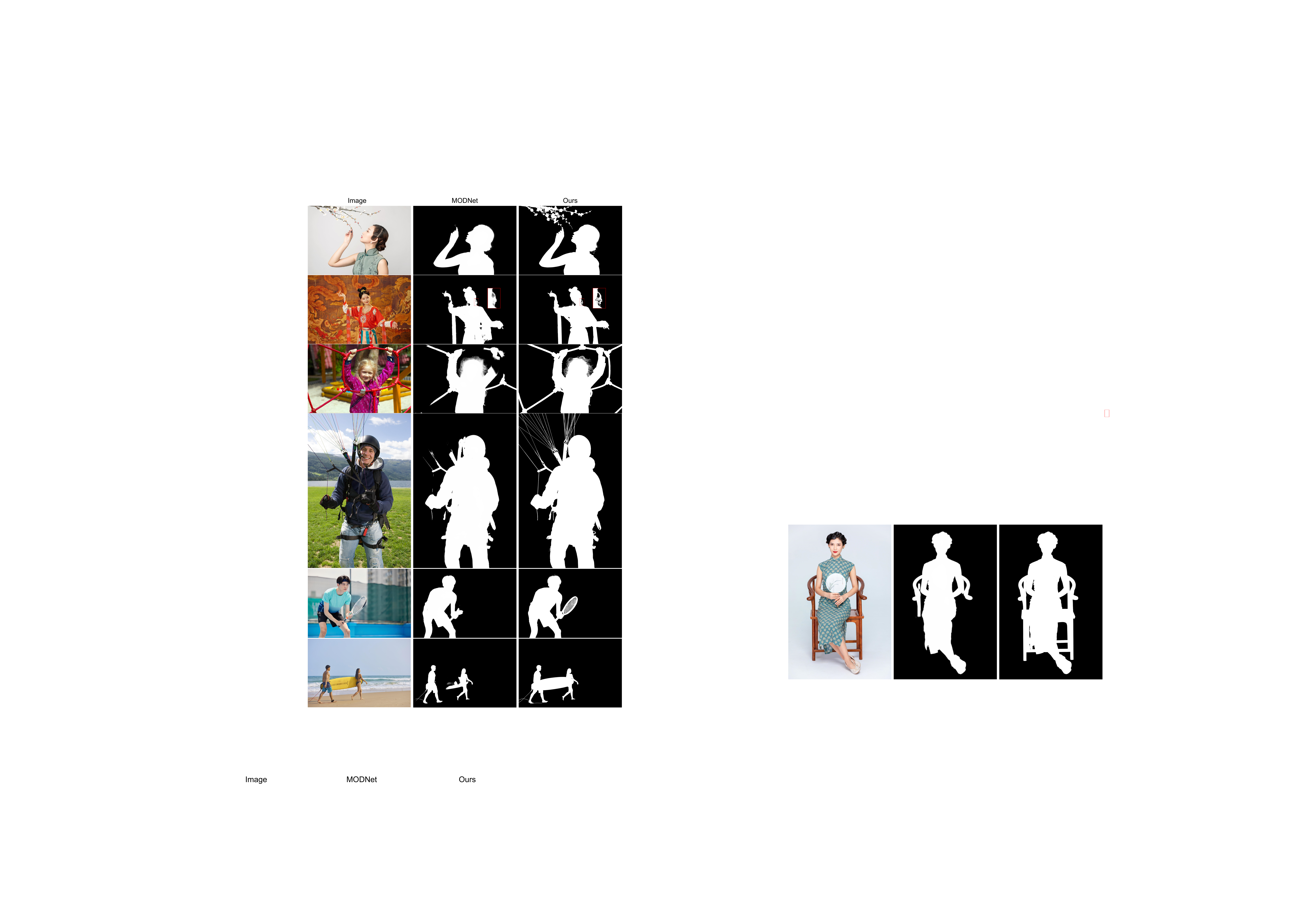}
  \caption{More visual comparisons between MODNet~\cite{forte2020f} and our SPG-IM on real world images.}
  %\Description{A woman and a girl in white dresses sit in an open car.}
  \label{fig:visual_real_1}
%\vspace{-10pt}
\end{figure*}

We display more representative visualizations on our Multi-Object-1K benchmark and real world images. Performance comparisons in Figure~\ref{fig:visual_multi_1}, ~\ref{fig:visual_multi_2}, and ~\ref{fig:visual_real_1} demonstrate the effectiveness and generalization of our situational perception guided image matting (SPG-IM), especially in the multi-object foreground scenes. Meanwhile, our SPG-IM can enhance both global semantic awareness and local details. The proposed Multi-Object-1K can further extend the image matting from the single-object foreground scenes to the complex multi-media situations.

\begin{table}[t]
\renewcommand\tabcolsep{3.5pt}
\begin{center}
\scalebox{1.0}{
\begin{tabular}{lcc}
\toprule
Method &Parameters (M) \qquad\qquad& Size (MB) \\
\midrule
LFM~\cite{zhang2019late}&225.9 &863.5 \\
%MODNet&6.5&25\\
GCA~\cite{li2020natural}&25.3 &96.6 \\
GCA$_{auto}$&80.0 & 305.3 \\
FBA~\cite{forte2020f}&34.7 & 138.8 \\
FBA$_{auto}$&89.4 & 347.5\\
MG$_{auto}$~\cite{yu2021mask}&84.4 &322.7 \\
\midrule
SPD&40.2 & 153.7\\
SPG-IM&61.2 & 234.1 \\
%\cmidrule(l){1-1}

\bottomrule
\end{tabular}}
\end{center}
%\vspace{-10pt}
\caption{Model size comparison. GCA$_{auto}$, FBA$_{auto}$, and MG$_{auto}$ use DeepLabV3+ with Xception backbone for the segmentation prior (automatic trimap or mask generation). The implementation of LFM~\cite{zhang2019late} network is based on the TensorFlow~\cite{abadi2016tensorflow} library.}
\label{model_size}
%\vspace{-18pt}
\end{table}

\clearpage

%\bibliographystyle{ACM-Reference-Format}
%\bibliography{sample-base}

%\appendix





\end{document}